\documentclass[10pt,twocolumn,letterpaper]{article}

\usepackage{cvpr}              

\usepackage{graphicx}
\usepackage{amsmath}
\usepackage{amssymb}
\usepackage{balance}
\usepackage{soul}
\usepackage{enumitem}
\usepackage{adjustbox}
\newcommand\tab[1][1cm]{\hspace*{#1}}

\usepackage[pagebackref,breaklinks,colorlinks]{hyperref}

\usepackage[capitalize]{cleveref}
\crefname{section}{Sec.}{Secs.}
\Crefname{section}{Section}{Sections}
\Crefname{table}{Table}{Tables}
\crefname{table}{Tab.}{Tabs.}

\def\cvprPaperID{10} 
\def\confName{CVPR}
\def\confYear{2022}

\begin{document}

\title{On the Exploitation of Deepfake Model Recognition}


\author{Luca Guarnera $^1$ \tab Oliver Giudice $^2$ \tab Matthias Nie\ss ner $^3$ \tab Sebastiano Battiato $^1$ \\
\small $^1$ Department of Mathematics and Computer Science, University of Catania, Italy\\
\small $^2$ Banca d'Italia, Applied Research Team, IT Dept., Rome, Italy\\
\small $^3$ Technical University of Munich, Germany\\
{\tt\small luca.guarnera@unict.it} \tab[0.5cm] {\tt\small giudice@dmi.unict.it} \tab[0.5cm] {\tt\small niessner@tum.de} \tab[0.5cm] {\tt\small battiato@dmi.unict.it}
}

\maketitle

\begin{abstract}

Despite recent advances in Generative Adversarial Networks (GANs), with special focus to the Deepfake phenomenon there is no a clear understanding neither in terms of explainability nor of recognition of the involved models. In particular, the recognition of a specific GAN model that generated the deepfake image compared to many other possible models created by the same generative architecture (e.g. StyleGAN) is a task not yet completely addressed in the state-of-the-art. In this work, a robust processing pipeline to evaluate the possibility to point-out analytic fingerprints for Deepfake model recognition is presented. After exploiting the latent space of 50 slightly different models through an in-depth analysis on the generated images, a proper encoder was trained to discriminate among these models obtaining a classification accuracy of over 96\%. Once demonstrated the possibility to discriminate extremely similar images, a dedicated metric exploiting the insights discovered in the latent space was introduced. By achieving a final accuracy of more than 94\% for the Model Recognition task on images generated by models not employed in the training phase, this study takes an important step in countering the Deepfake phenomenon introducing a sort of signature in some sense similar to those employed in the multimedia forensics field (e.g. for camera source identification task, image ballistics task, etc).

\end{abstract}

\section{Introduction}
The phenomenon of the so-called Deepfake has been observed and treated with growing concerns and interests in recent years by the research community. In summary, there is quite agreement on the nature and disruptive potential of this technology: based on a particular application of  Generative Adversarial Networks (GAN)~\cite{goodfellow2014generative} they synthesize extremely realistic multimedia content. 

Legal literature in recent years has dealt with the subject from the perspective of national security and safeguarding of the democratic order and the protection of privacy, as well as the reputational damage of natural and legal persons, as well as the crime related to non-consensual dissemination of sexually explicit images. For this reason, the computer science community has addressed the problem by focusing on proving the authenticity of an image \cite{hasan2019combating,ong2021protecting} or by developing more and more sophisticated Deepfake Detection techniques \cite{verdoliva2020media,mirsky2021creation,tolosana2020deepfakes,masood2021deepfakes,swathi2021deepfake,guarnera2020deepfake}. 

Deepfake images are so realistic that even specialists can find it difficult to detect them \cite{hulzebosch2020detecting}. On the other hand, it seems that state-of-the-art (SOTA) techniques are surprisingly efficient to spot them and even to recognize the architecture. Thus, a new and more difficult challenge could be to recognize the model (defined as the set of weights/parameters) which is an instance of a specific architecture. Having a Model Recognition solution could enable the possibility of attributing an image to a specific model owner (i.e. for Intellectual Property). 


Detection and model recognition are definitely required to approach and counter the Deepfake phenomenon. To prove the ownership or authenticity of an image generated by a given model of a specific architecture, novel strategies with proper ``metrics" are needed. This is of utmost importance in the media forensics field, to reconstruct the origin of an image~\cite{farid2008digital}. New definitions can be given for the image ballistics investigation when Deepfake images are involved: SOTA Deepfake Detectors and Architecture Classifiers could be associated with the camera model identification task whilst the camera source identification task could be equivalent to a Deepfake Model Recognition task, given a specific architecture. This last task has not yet been investigated in literature.

\begin{figure*}[t!]
    \centering
    \includegraphics[width=\textwidth]{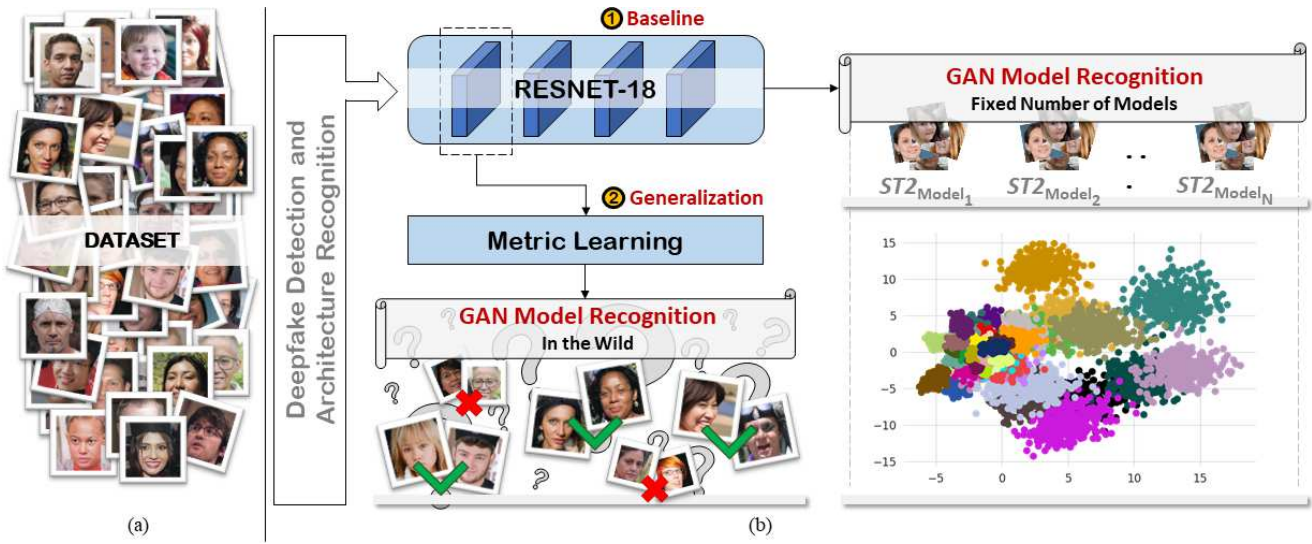}
    \caption{Schematic representation of the robust processing pipeline proposed for Deepfake Model Recognition. At first, a dataset has to be created (a) by fine-tuning different models and thus generating images. In general, an efficient Deepfake detector/classifier could be employed at first from those available in literature. Then, an encoder (b) is trained for the recognition task on the a-priori known models. The obtained feature vectors on a two-dimensional space shows the discriminative power of features encoded by the baseline encoder. Finally, the first layer of the trained encoder is exploited in a metric learning approach.}
   \label{fig:proposedapproach}
\end{figure*}

A fundamental element of all GAN architectures is their latent space. In summary, the N-dimensional latent space of a GAN represents all the patterns learned by the ``Generator" in the training phase related to the training data. Essentially, it maps the space to specific features of the training data. This is fundamentally used to re-create synthetic multimedia data with features similar to those of the dataset used during the training phase. Each time a GAN model is trained the latent space will be different. Several attempts to introduce GAN explainability have been proposed by exploiting the visual concept vocabulary~\cite{schwettmann2021toward}, analyzing/basing on node collapse~\cite{bau2019seeing} , and by interpreting the latent semantic space ~\cite{shen2020interpreting}.
StyleGAN \cite{abdal2019image2stylegan}, among others, is an example of generative architecture that introduced a specific latent space to incorporate semantic information for face editing. 

In this paper, the latent spaces of 50 slightly different StyleGAN2~\cite{karras2020analyzing} models were explored in depth.  To this end, to reduce any correlation to high-level semantic content, only images of faces were taken into account. First, 50 different generative models were trained and the generated images were analyzed considering standard visual metrics and distances with respect to the latent space of each model. The discovered insights, confirmed the existence of traces linked to the StyleGAN2 architecture, thus explaining why SOTA detection solutions are able to obtain astonishing results. As a matter of fact, the Deepfake Detection task is simpler than recognizing a model which generated an image from a given architecture. To propose a novel processing pipeline and methodology for this new and more challenging task, a Resnet-18~\cite{he2016deep} encoder was trained with the objective of classifying images among 50 a-priori known slightly different models of the StyleGAN2 architecture. An overall accuracy of $96.24\%$ demonstrated the possibility of discriminating extremely similar images for the Deepfake Model Recognition task. Thus, a dedicated metric exploiting the encoded feature was introduced. Experimental results demonstrated the effectiveness of the proposed metric, for making the first milestone in countering the Deepfake phenomenon by introducing a method equivalent to the camera source identification task which is well known to image forensics experts. 
Finally, the processing pipeline and metric were experimentally tested to understand their limitations and generalizing ability. The main contributions of this paper are summarized as follows:%
\begin{enumerate}
    \item An in-depth exploration of the latent space of the slightly different StyleGAN2 models, and a description of the insights discovered;
    \item A new pipeline for the Deepfake Model Recognition task, given a specific architecture. Figure~\ref{fig:proposedapproach} schematically shows the overall framework;
    \item A methodology to encode a robust descriptor and a respective metric for Deepfake Model Recognition. 
    \item Datasets and code for Deepfake Model creation and recognition, publicly available.
\end{enumerate}

The remainder of this paper is organized as follows: Section \ref{sec:related} briefly overviews related works from literature. Section \ref{sec:dataset} introduces the notation and a detailed description of the methodology used to train different models and generate images. Then, Section \ref{sec:method} presents the pipeline for Deefake Model Recognition which is finally tested and discussed in Section \ref{sec:results}. Finally, a summarized discussion is reported in Section \ref{sec:conclusion} that concludes the paper with hints for future works.


\section{Related Works}
\label{sec:related}

Deepfake images are generally created by techniques based on GANs, originally introduced by Goodfellow et al.~\cite{goodfellow2014generative}. Since 2014, many architectures and variants have been proposed \cite{creswell2018generative}; one of the most recent and powerful methods regarding the entire-face synthesis is the Style Generative Adversarial Network architecture or commonly called \mbox{StyleGAN \cite{karras2019style}}. It is capable not only of generating impressive photo-realistic and high-quality faces, but also offers some kind of control in terms of the overall style of the generated image at different levels of detail. Later, Karras et al. made some improvements to the technique proposing the StyleGAN2~\cite{karras2020analyzing} architecture. Images generated by StyleGAN2 are extremely realistic and many techniques have already been proposed to detect their fakeness. On this topic, several overviews on Media forensics with a particular focus on Deepfakes have been recently published~\cite{tolosana2020deepfakes, verdoliva2020media,mirsky2021creation,frank2020leveraging,zhang2019detecting}.

To analyze the authenticity of a digital media (image, audio, video) and reconstruct its history (provenance and phylogeny), SOTA techniques could be divided into two major categories: active and passive~\cite{piva2013overview,battiato2016multimedia}.

Active techniques embed an hidden signature on images by varying the image creation process (generative process when dealing with Deepfakes). This signature  can be detected at any time in the life-cycle of an image. Active techniques related to Deepfake images have been proposed recently: watermarking techniques \cite{vukotic2018deep,zhang2019invisible}, which embed information on pixels and Neural Network Watermarking techniques \cite{zhang2018protecting,uchida2017embedding,yu2020responsible}, which embed information on the network parameters. While being highly efficient, if a watermark is not embedded preliminarly on images only passive techniques could be employed. Moreover, if the embedding process is known, watermark canceling and extraction could be performed.

However, passive techniques attempt to extract hidden signatures accidentally left by the creation process of images. These techniques are less efficient than active ones but they could be employed in any scenario. In general, they can only deal with easier tasks, such as: Deepfake Detection, which is a binary classification task between real and fake images; and Deepfake Attribution (or GAN architecture recognition), which is a multi-class classification task. Indeed, Deepfake Detection techniques in SOTA seem to solve an extremely easy task~\cite{wang2020cnn}. This fact was demonstrated to be related to a sort of fingerprint, a convolutional trace, left on the images by the GAN generative process~\cite{guarnera2020fighting,marra2019gans}. The underlying trace, invisible to the human eye, can be easily spotted in the frequency domain~\cite{durall2020watch}: Discrete Cosine Transform (DCT) analysis, for example, could lead to the development of extremely powerful, explainable and efficient detectors~\cite{giudice2021fighting,durall2019unmasking}. Many techniques are available in literature that focus on different architectures and scenarios \cite{goebel2021detection,yu2019attributing,dang2020detection,agarwal2020detecting,wang2019fakespotter,guarnera2020fighting}.

All the aforementioned passive approaches are able to achieve high accuracy scores, even in the GAN Architecture Classification task. 
Creation, detection, and architecture recognition have been the main topics addressed by the scientific community 
A first study on the recognition task was recently proposed by Asnani et al.~\cite{asnani2021reverse}: an “agnostic” prediction solution of the architecture that generated a given image. For instance, given an image, they are able to infer the structure (with an overall hyper-parameter setting) of the GAN architecture which generated that image. In particular, they make use of a specific single instance of a model for each considered architecture.

None of the above-mentioned SOTA approaches (Deepfake Detection, GAN Architecture Classification, Attribute Recognition) are sufficient to recognize if an image was generated by someone employing not only a given architecture but a specific model of that given architecture, defined by its set of unique (hyper) parameters~\cite{zhang2020not} (among different instances). This last task could be defined as Deepfake Model Recognition and it has not yet been completely addressed in literature. This paper deals with this lack in literature by proposing a preliminary metric-based solution and methodology.

\begin{figure*}[t!]
    \centering
    \includegraphics[width=\textwidth]{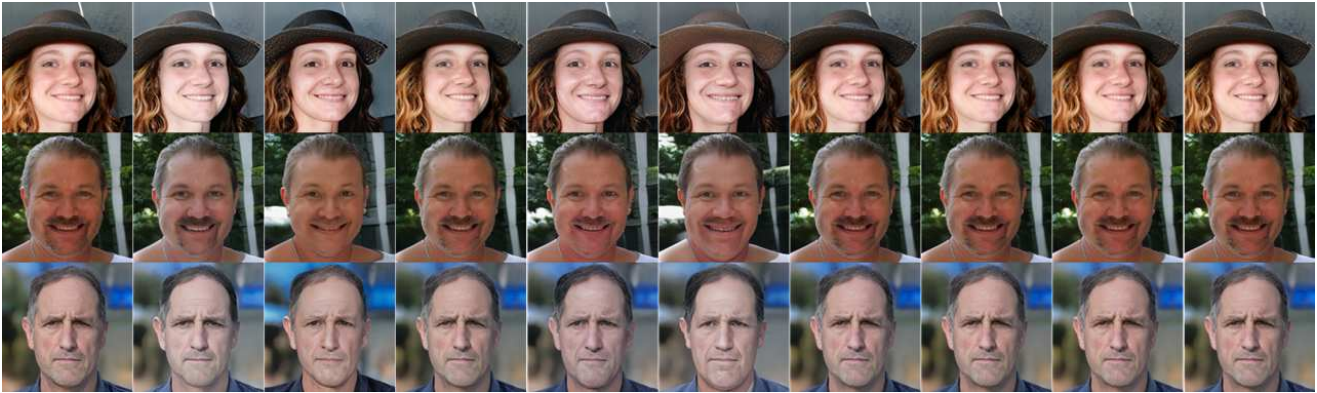}
    \caption{Generated image examples from the slightly different models employed in the experiments.} 

   \label{fig:datasets}
\end{figure*}

\vspace{-0.1cm}

\section{A Dataset for Model Recognition}
\label{sec:dataset}

StyleGAN2 is the architecture considered as the first example of delving into the Deepfake Model Recognition task. Specifically,  StyleGAN2-ADA~\cite{NEURIPS2020_8d30aa96} was the employed implementation given its fine-tuning training easiness which, with few training examples, is able to obtain almost completely artifact-free Deepfake face images. Moreover, the StyleGAN2-ADA implementation easily allows to freeze the weights/parameters of a user-defined number of layers in the ``Discriminator" during the training operations. In order to have this freezing ability also available for the layers of the ``Generator", the original implementation of StyleGAN2-ADA was customized. This was of key importance for the task of training different models with slightly different parameters (aka weights) from which generating images extremely similar to each other and almost ``identical" to human eyes.
To facilitate the reading of this paper, given the high number of slightly different models and the complexity of the created dataset, a mathematical formalism will be introduced in the following.



\color{red}





\color{black}





\color{black}

\subsection{The Models}
\label{sec:dataset-creation}

\begin{table}[t!]
    \centering
    \begin{tabular}{|c|c|}
    \hline
    \textbf{Generator}    & \textbf{Discriminator} \\ \hline
    synthesis.b1024.conv0 & b4.mbstd               \\
    synthesis.b1024.conv1 & b4.conv                \\
    synthesis.b1024.torgb & b4.fc                  \\
    synthesis.b1024:0     & b4.out                 \\
    synthesis.b1024:1     &                        \\ \hline
    \end{tabular}
    \caption{StyleGAN2 Generator $G$ and Discriminator $D$ structure. The first column describes layer $b1024$ in the synthesis block in $G$; the second column represents the structure of layer $b4$ in $D$. These blocks represent the layers that were re-trained during the fine-tuning procedure}
    \label{tab:GDlayers}
\end{table}



Let $M_{ST2}$ be the well-known pre-trained StyleGAN2 model~\footnote{\url{https://github.com/NVlabs/stylegan2}}, several fine-tune  trainings were carried out by freezing all the layers except the latter in the Generator ($G$) and Disciminator ($D$) considering the FFHQ as the training dataset (employed in the same way as described in the official StyleGAN2-ADA Github repository~\footnote{\url{https://github.com/NVlabs/stylegan2-ada-pytorch}}). Training operations were carried out by varying only the following hyper-parameters: \textit{k-img} $\in \{1,2, ... ,10\}$ and $p \in [0, 1]$, which describe training-set augmentation probability. Note that these hyper-parameters affect the training process only, while all the other hyper-parameters were left as StyleGAN2-ADA defaults. All possible combinations of pairs $(\bar k, \bar p)$ with $\bar k \in$ \textit{k-img} and $\bar p \in p$  were considered (examples (1, 0.1), (1, 0.2), ... (10, 0.9), (10, 1.0)) to obtain a total of $100$ StyleGAN2-ADA fine-tuned models ($M_{ST2-ADA}$). For each $M_{ST2-ADA}$, $1,000$ images were generated for a total of $100,000$ images.

Figure~\ref{fig:datasets} shows samples of the generated images: the first column shows images created by the pre-trained StyleGAN2 model ($M_{ST2}$); the remaining columns represent samples of the images generated by the fine-tuned StyleGAN2-ADA models, with proper combinations of ($\bar k$, $\bar p$) parameters.

\begin{figure*}[t!]
    \centering
    \includegraphics[width=14cm]{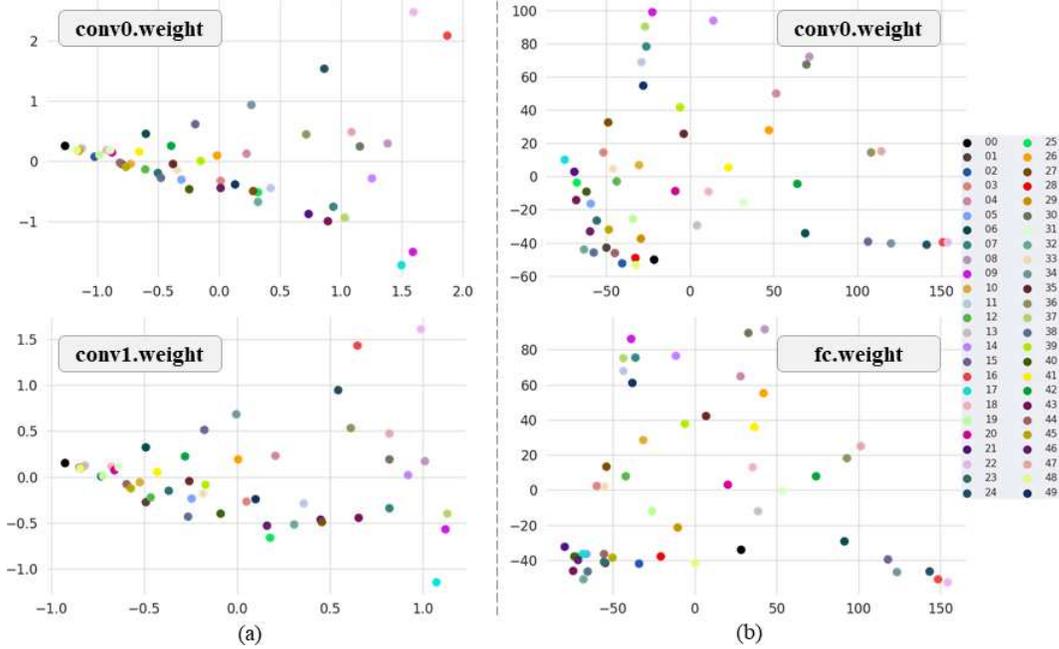}
    \caption{2D PCA of the parameters extracted from the last layer of the Generator (a) and the Discriminator (b) of each of the $50$ models.} 
   \label{fig:param}
\end{figure*}

Let $M_i(W)$ be the \textit{i-th} model of $M_{ST2-ADA}$ uniquely defined by its set of parameters (weights) $W=(W_G,W_D)$ with $W_G=\{w_{G1},w_{G2}, ... , w_{Gn}\}$ and $W_D=\{w_{D1},w_{D2}, ... , w_{Dm}\}$; $n$ and $m$ are the total number of layers in $G$ and $D$ respectively (e.g., $w_{G1}$ represents all weights of layer1 in $G$, while $w_{G2}$ all weights of layer2 in $G$, etc.). It is worth to highlight that each model $M_i(W)$ is uniquely defined by the values of its parameters $W$ and thus creates a unique latent space. The $W$ of all $M_{ST2-ADA}$-created models will differ only by $w_{Gn} \in W_G$ and $w_{Dm} \in W_D$ which correspond to weights belonging only to the last layer of $G$ and $D$. Specifically, as far as the ``Generator" is concerned, only the layer named $b1024$ was not frozen. This layer, represents the last image synthesis operation and directly affects the image creation process. Similarly for the ``Discriminator" only the layer named $b4$ was not frozen. A detailed description of the composition of the mentioned layers, for both the Generator and Discriminator, is reported in Table~\ref{tab:GDlayers}. The described training process produced slight differences in each model parameter set $W$. This is demonstrated in Figure~\ref{fig:param} where a plot obtained from a 2-dimensional PCA reduction of the weights extracted from the last layer of the Generator and the Discriminator of each $M_i(W)$ is reported showing an initial exploration of the corresponding latent space.

\subsection{The images}
\label{sec:dataset-analysis}

Let $I_{M_i(W)}$ be the set of images generated by the model $M_i(W)$, the  $I_{M_{ST2-ADA}(W)}$ images will turn out to be very similar to each other in terms of visual appearance. This is due to the slight differences in the weights $W$ among the different models $M_i(W)$.

\begin{figure*}[t!]
    \centering
    \includegraphics[width=13cm]{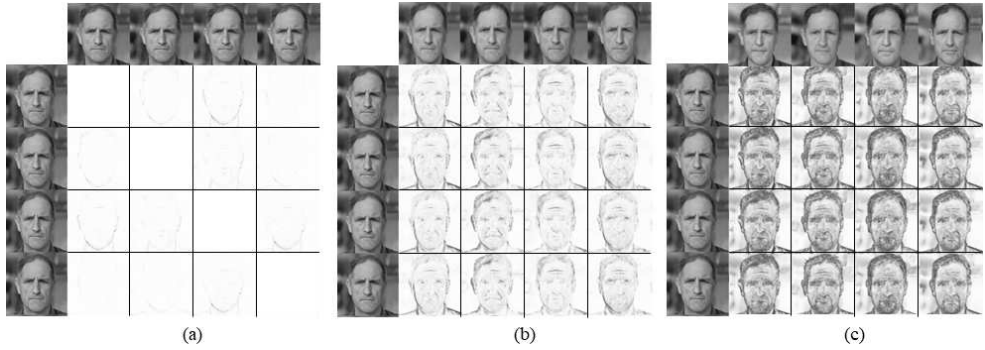}
    \caption{SSIM maps comparisons between gray-scale images. (a) shows samples of extremely similar images; (b) shows samples where the differences are related mainly to the smoothing effects. (c) shows examples where the difference is more evident and related to misalignments.}
   \label{fig:SSIM}
\end{figure*}

In Figure~\ref{fig:datasets} are reported several images of $I_{M_i(W)}$: some may show visible artifacts, whereas others may present totally imperceptible variations, not only w.r.t. the basic StyleGAN2 model $M_{ST2}$ (Figure~\ref{fig:datasets} first column) but also w.r.t. all the other models  $M_{ST2-ADA}$ (Figure~\ref{fig:datasets} from column 2 onwards).
In order to understand these imperceptible differences, the \textit{Structural Similarity Index Measure} (SSIM) was employed between the image samples generated by $M_{ST2}$ vs. all those generated by the $M_{ST2-ADA}$ models. Figure~\ref{fig:SSIM} shows an example of the results obtained in form of a matrix comparison. It is possible to note that in almost all cases the variations between each pair of images are very small (Figure~\ref{fig:SSIM}(a)) consequently the Model Recognition could be a very complicated task. 


\subsection{Training and Test data definition}

In the previous Sections, the creation of $M_i(W)$ with $i=1, \dots, 100$  different models were described. Only 50 ($M_{1-50}$) of the 100 $M_i(W)$ models were randomly selected for the training procedures of the algorithms presented in the following Sections. Specifically, for each of the selected model, $1,000$ images were generated.  The ``baseline" and the ``generalization" phases of the pipeline, as shown in Figure~\ref{fig:proposedapproach}, were trained on this set of images.
Testing was carried out on different sets of images: the ``baseline" phase was tested on $300$ other images generated by $M_{1-50}$ models and the ``generalization" phase was tested to compute proper scores on $300$ images generated by each $M_{51-100}$ models (for a total of $30,000$ images of the two sets).

\begin{figure*}[t!]
    \centering
    \includegraphics[width=13cm]{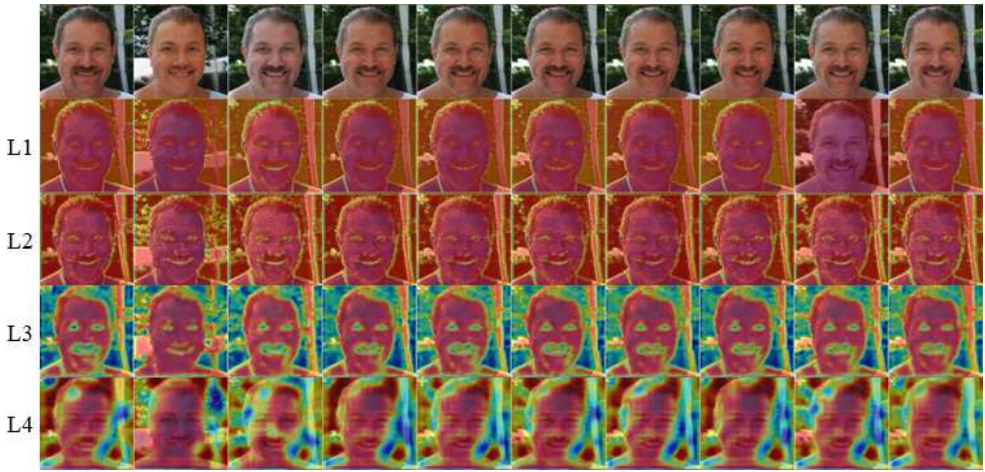}
    \caption{Attention maps of the first four layers (L1, L2, L3, L4) of the Resnet-18 trained for the a-priori known model classification task.} 
   \label{fig:gradcam}
\end{figure*}

\section{Deepfake Model Recognition}
\label{sec:method}

The image dataset and models described in Section \ref{sec:dataset} only show the faces of people as content. This creates an extremely challenging scenario in which the characteristics can be summarized as follows: (i) there are images representing almost the same faces among the slightly different employed models; (ii) each face has very small variations depending on the model. These properties could be of great importance in obtaining invariance to high-level semantic features related to the image content. Thus, it is possible to build a robust processing pipeline that is able to capture only the discriminating parameters in the latent space of the $M_{ST2-ADA}$ models, making it not related to a specific face feature.

The objective of a Deepfake Model Recognition task is to assess whether two images were generated by the same model by defining a metric $L$. Formally,

\begin{equation}
    L(F(I^i_a), F(I^j_b)) \in [-1,1]
    \label{eq:L_metric_formula}
\end{equation}

where $I^i_a \in I_{M_i(W)}$ and $I^j_b \in I_{M_j(W)}$ are two images generated respectively by $M_i(W)$ and $M_j(W)$ models ($M_i \in M_{ST2-ADA}$ and $M_j \in M_{ST2-ADA}$), and $F(\cdot)$ is a function for fingerprint extraction from a given input image. The more $L$ is near to ``$1$", the higher is the probability that the two images were generated using the same model. On the other hand, $L$ near to ``$-1$", means that images were generated by different models.


The elements of Equation \ref{eq:L_metric_formula} are found in three different (but consequent) moments:
\begin{enumerate}
    \item To define and to train an encoder ($F$), able to discriminate 50 different classes ($M_{1-50}$) and demonstrate its effectiveness by measuring the accuracy even in noisy scenarios;
    \item To employ the trained classifier, as an embedding method ($F$) that is able to exploit the latent spaces, even for images generated by models never seen during the training phase ($M_{51-100}$);
    \item To define and to learn a robust metric $L$ to compare two images $I$ and assess whether they were generated by the same model.
\end{enumerate}

Figure~\ref{fig:proposedapproach} graphically schematizes the proposed approach: starting from the dataset generation phase to the final metric definition; it is possible to build a Model Recognition pipeline by exploiting a robust ``baseline" classifier and a metric for ``generalization", given a Deepfake architecture.


\subsection{Building an encoder for classification}
\label{subsec:RES}

\begin{figure*}[t!]
{\captionsetup{position=bottom,justification=centering}
     \subfloat[\label{subfig:accloss1}]{%
       \includegraphics[width=0.3\linewidth]{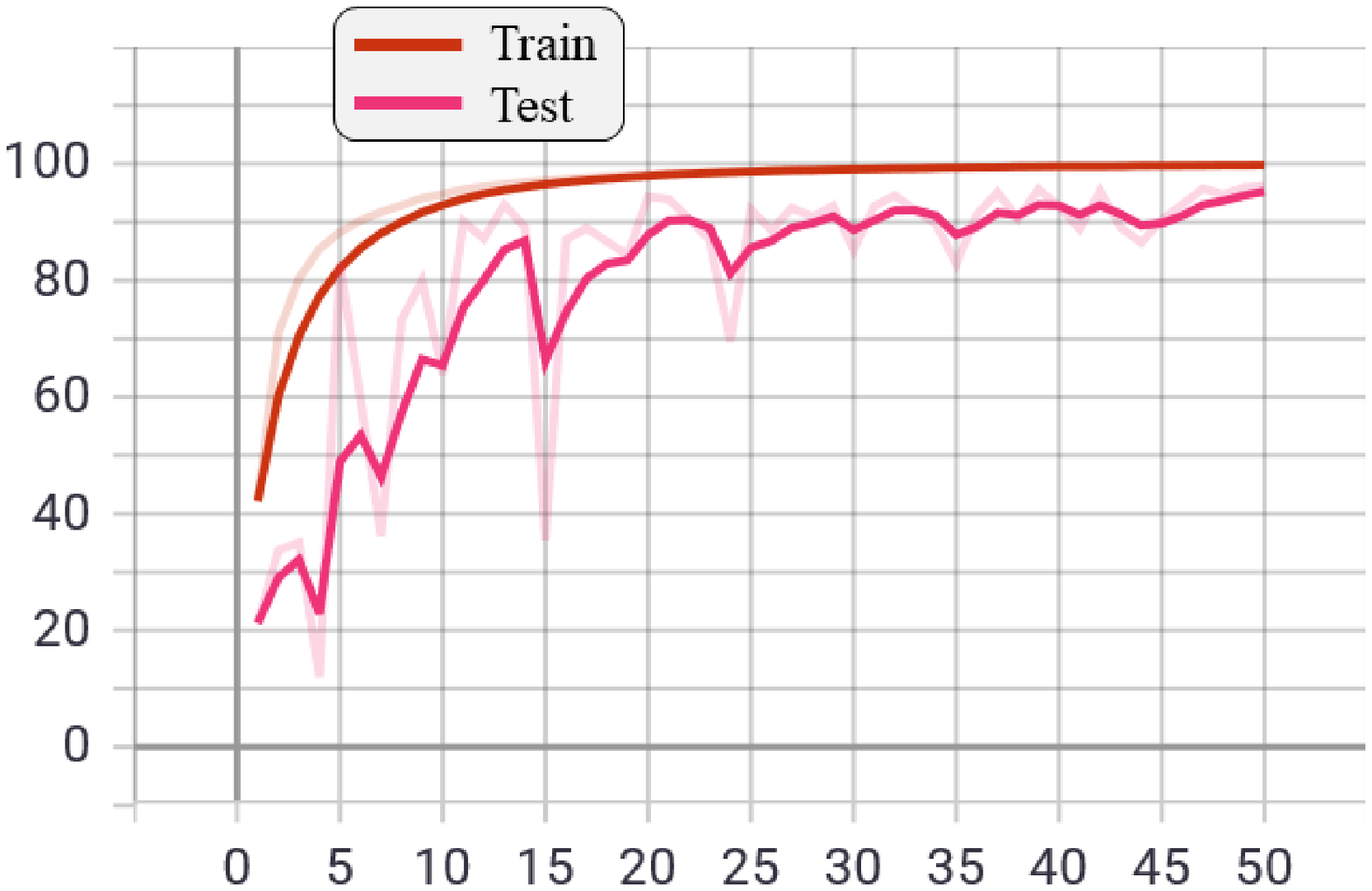}
     }
     \hfill
     \subfloat[\label{subfig:accloss2}]{%
       \includegraphics[width=0.3\linewidth]{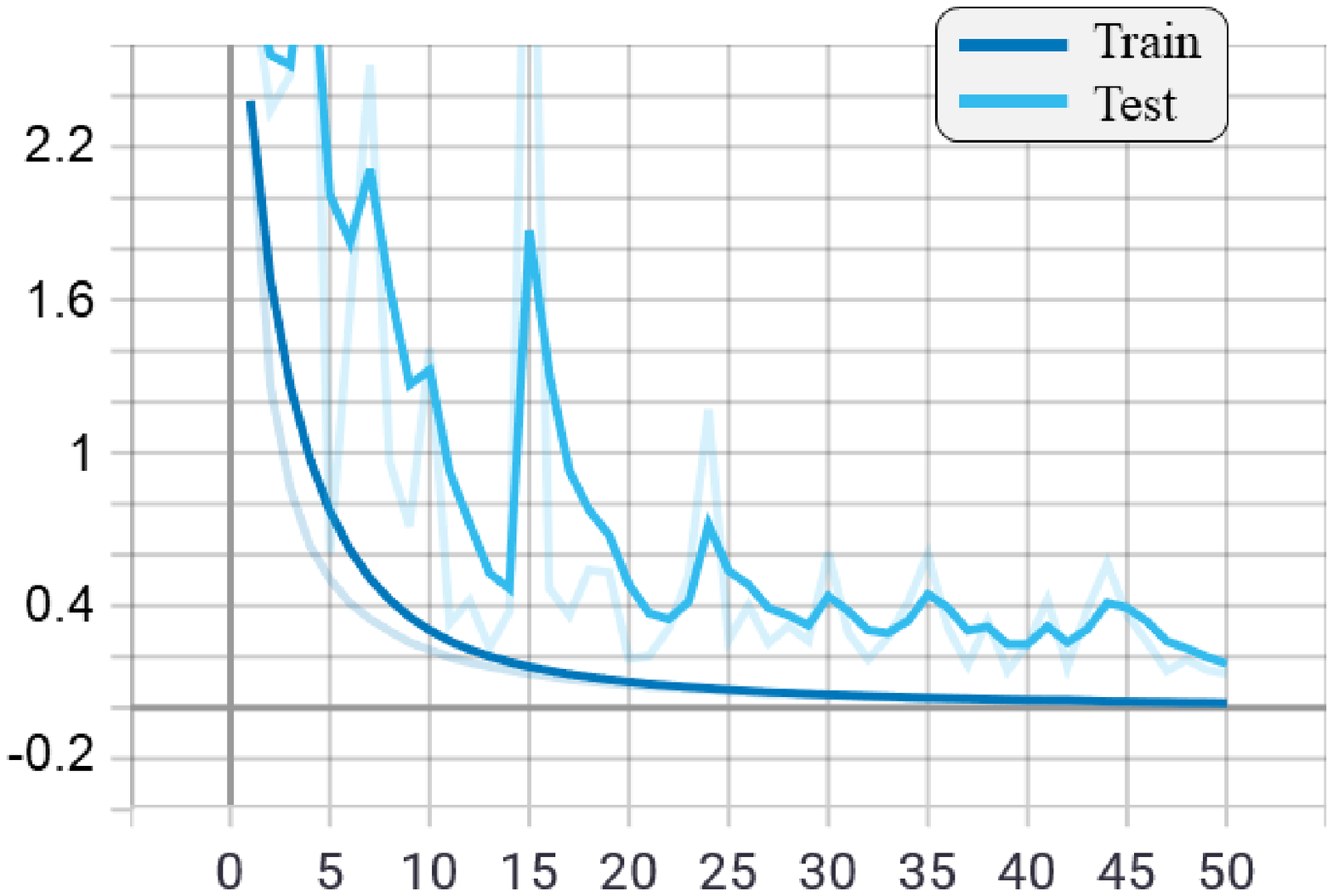}
     }
    \hfill
     \subfloat[\label{subfig:confusion1}]{%
       \includegraphics[width=0.22\linewidth]{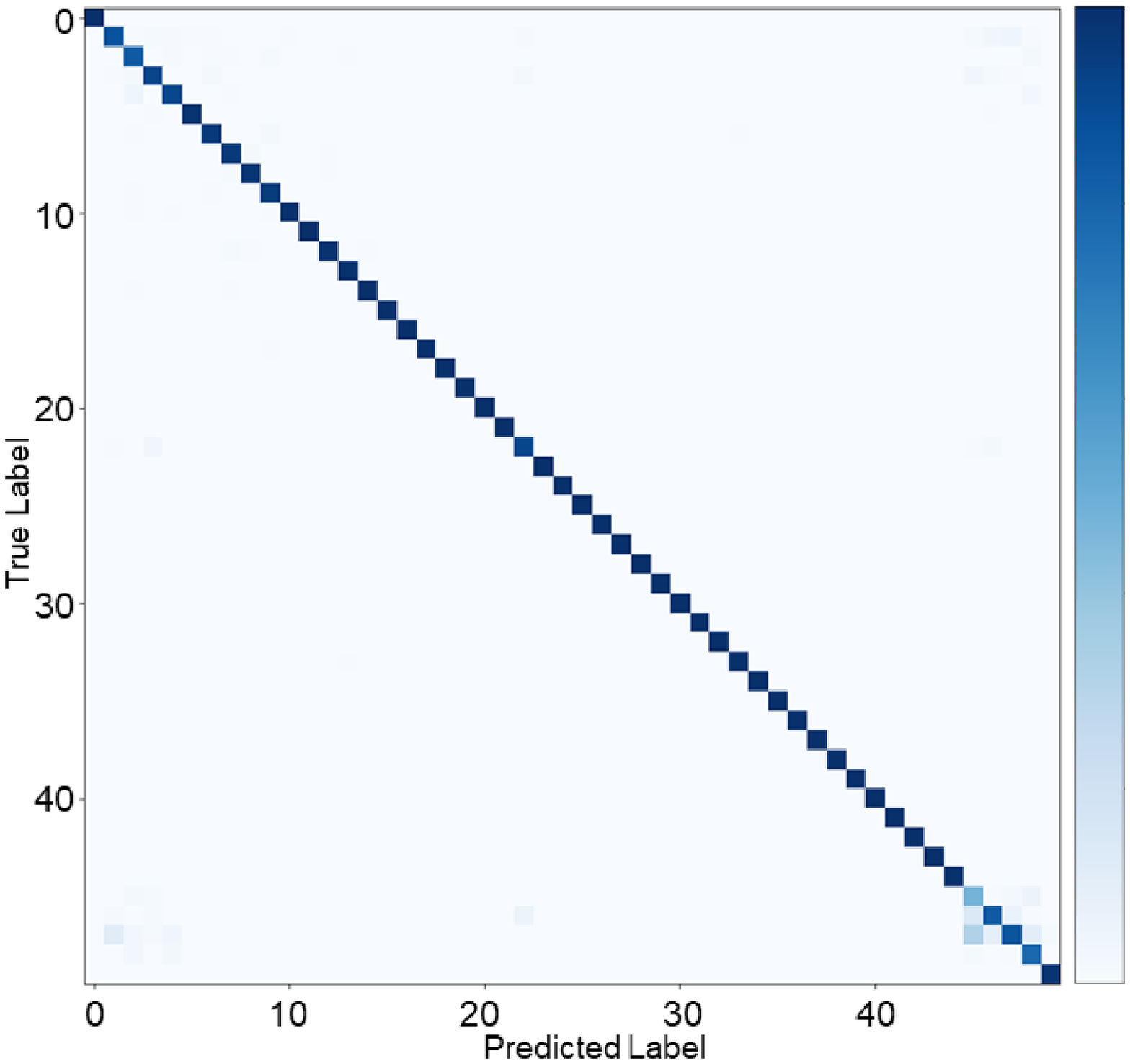}
     }}
    \hfill
     \caption{Training accuracy (a) and loss (b) through 50 epochs on raw-images. The confusion matrix in (c) shows classification results obtained by the best classification model, over 50 classes.}
     \label{fig:AccLossRESNET}
\end{figure*}

The Resnet-18 encoder was employed for the classification task of the images generated by $M_{1-50}$. Specifically, the Pytorch implementation of Resnet was used starting from the pre-trained version trained on ImageNet (publicly available in the torchvision module).
 A fully-connected layer with an output size of $50$ followed by a SoftMax were added to the last layer of the Resnet-18 in order to be trained on the specific task. The following settings were employed for training: learning rate = $0.001$, batch size = $30$, epochs = $50$, ``cross entropy loss" as loss function and the Stochastic Gradient Descent (with momentum = 0.9). Figure~\ref{fig:AccLossRESNET} shows the accuracy (a) and loss (b) values of the training phase. 

In order to define the best $F$($\cdot$) (see Equation~\ref{eq:L_metric_formula}), an evaluation on images was carried out with respect to attention heat-maps~\cite{selvaraju2017grad} for each layer of the trained Resnet-18. While Layer-1 encodes features as far as possible both from the final task and from the semantics, it is possible to note that Layer-4 concentrates much more on face skin and key-points than on the overall image parts (Figure~\ref{fig:gradcam}). Thus, the best $F$($\cdot$) would be the Layer-1. Moreover, Layer-1 is the smallest in terms of dimensions and would be better for computational costs for any kind of further processing. We also note that all parameters of the employed Resnet were trained to solve the proposed task. This implies that even the first layers are able to extract salient features. 

\begin{figure*}[t!]
{\captionsetup{position=bottom,justification=centering}
     \subfloat[\label{subfig:accloss3}]{%
       \includegraphics[width=0.3\linewidth]{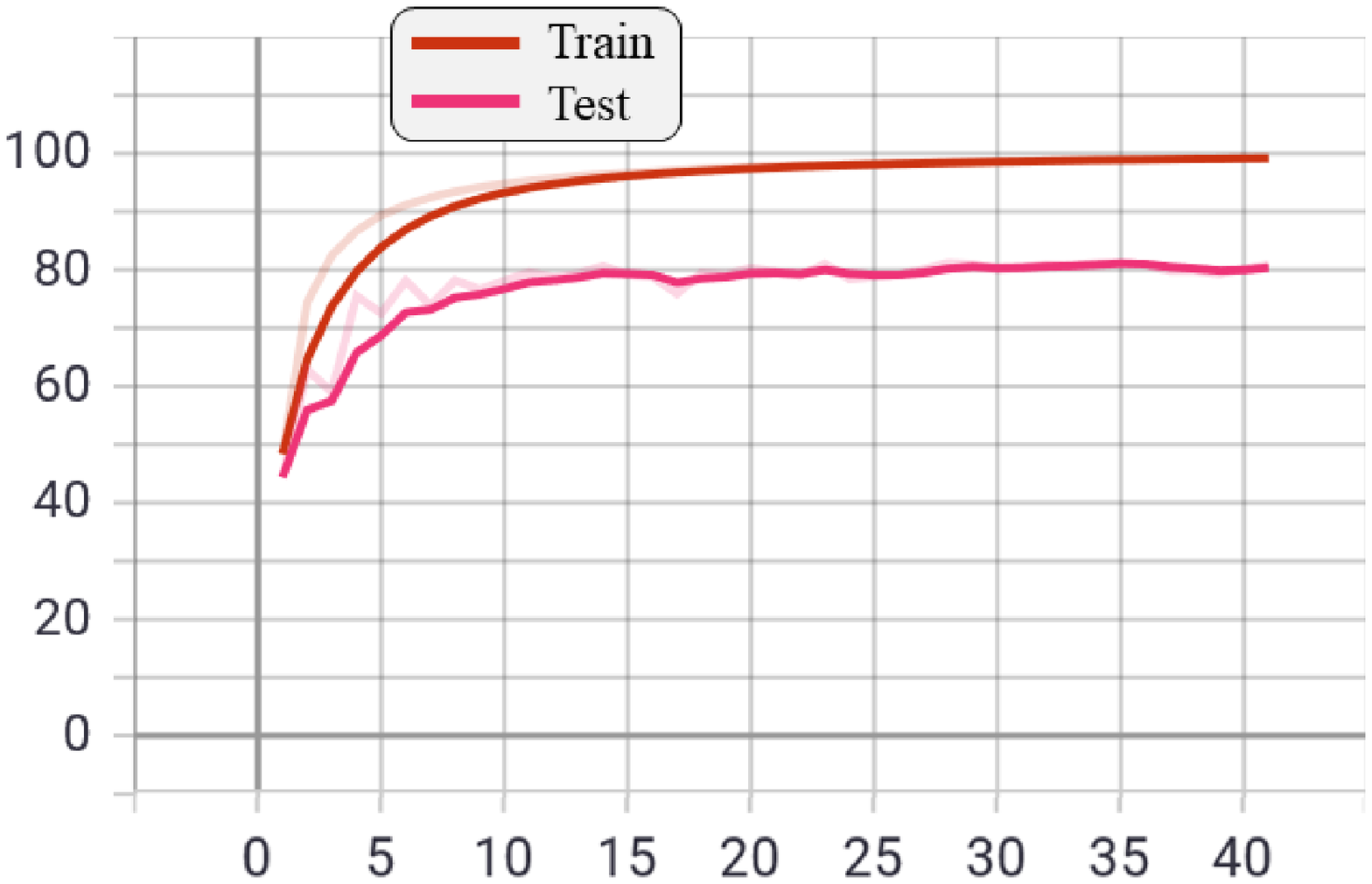}
     }
     \hfill
     \subfloat[\label{subfig:accloss4}]{%
       \includegraphics[width=0.3\linewidth]{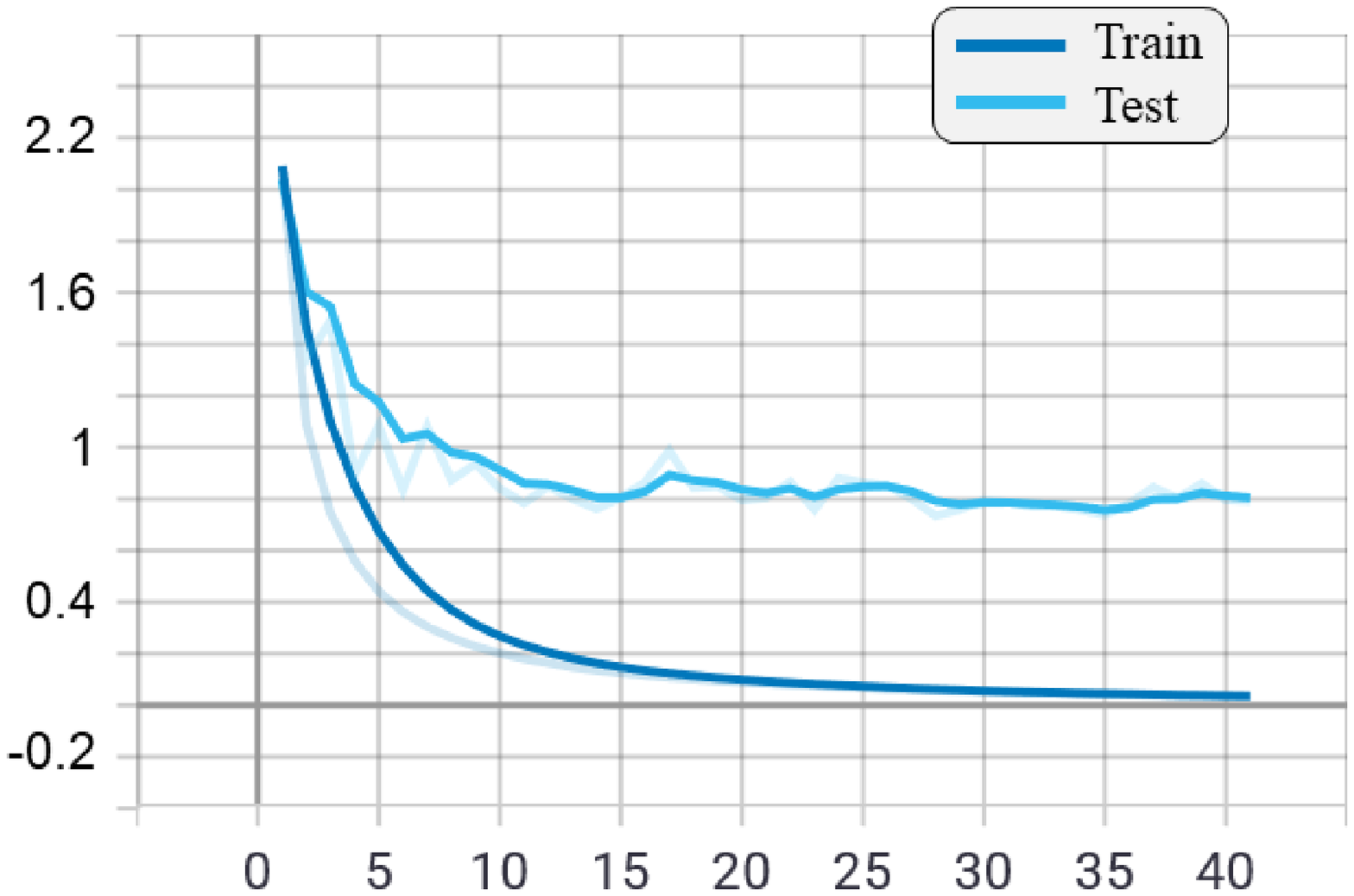}
     }
    \hfill
     \subfloat[\label{subfig:confusione2}]{%
       \includegraphics[width=0.22\linewidth]{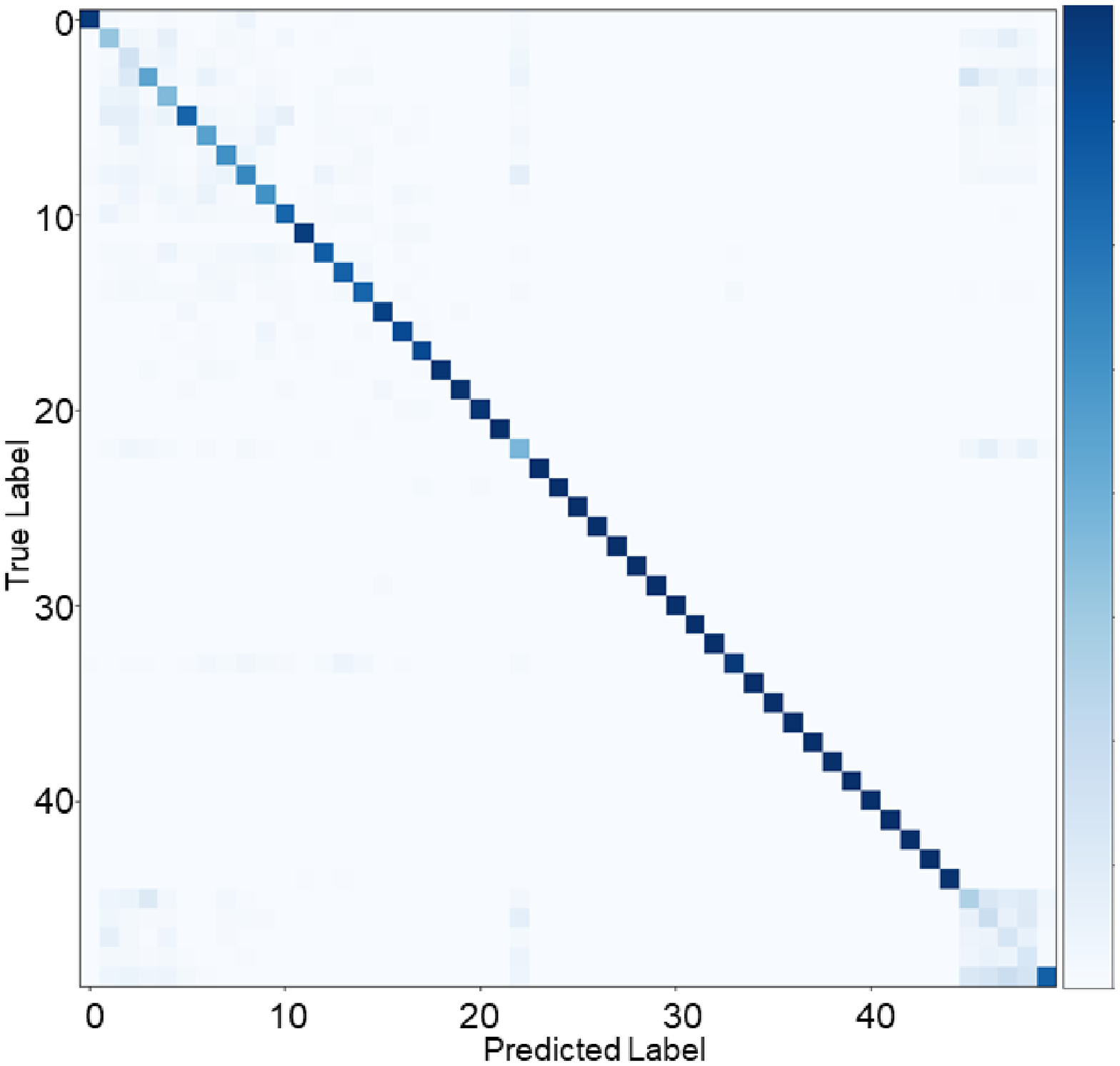}
     }}
    \hfill
     \caption{Training accuracy (a) and loss (b) through 41 epochs on noisy images. The confusion matrix in (c) shows classification results obtained by the best classification model, over 50 classes.}
     \label{fig:AccLossRESNETattacks}
\end{figure*}

\subsection{Learning a metric}
\label{subsec:metric}

The best model obtained in the operations described in Section~\ref{subsec:RES} was selected as $F$ w.r.t. Equation \ref{eq:L_metric_formula}. Specifically, given a generic image $I$ created from any model $M_{ST2-ADA}$, given the feature vector extracted from the output of Layer-1 of the trained Resnet-18, a tensor with dimensions $(1, 64, 256, 256)$, $F(I)$ is the function that extracts this tensor, flattens it and then applies a Singular Value Decomposition (SVD). SVD was fitted on $50\%$ of the training-set samples. The explained variance was employed as a score to find the best reduction ($400$, with a corresponding explained variance of $74\%$).

Given $F$, the definition of $L$ will be carried out by means of a metric learning approach developed specifically for working with high-dimensional data~\cite{liu2012metric}. This approach was chosen for being able to handle a specific kind of input feature vector and for being a weakly supervised metric learning approach, which is the case, given the weak quality of labelling limited only to similar or dissimilar labels. 

\color{black}

\color{black}

\section{Experimental Results}
\label{sec:results}
The solution described in the previous Sections and summarized in Figure~\ref{fig:proposedapproach} was deeply analyzed and tested on images belonging to the data described in Section~\ref{sec:dataset}. At first the trained feature extraction method ($F$) will be demonstrated to be robust and efficient in discriminating the images generated by 50 slightly different models. Moreover, $F$ is tested for robustness on a noisy version of the images on which different attacks were applied (Section~\ref{sub:resultsResnet18}). Finally, a metric $L$ is defined and experimentally tested in its ability to overcome the state-of-the-art and be the first solution for Deepfake Model Recognition (Section~\ref{sub:resultsmetric}).


\subsection{A-priori known models classification results}
\label{sub:resultsResnet18}
Given a set of known models of a given architecture ($M_{1-50}$), and the respective $1,000 \times 50$ images generated from them, a training-validation set split was performed (75\%-25\%). A Resnet-18 architecture was trained on the training-set and the training/validation loss plots obtained are shown in Figure \ref{fig:AccLossRESNET} (a) and (b). Then, the trained model was evaluated in the classification task on the validation set. Results, are shown in Figure \ref{fig:AccLossRESNET} in the form of a heat-map confusion matrix (c). An average classification accuracy value of $96.24\%$ was obtained demonstrating the effectiveness of the solution on a-priori known models.

Robustness evaluations were carried out by training a new Resnet-18 model on images where several filters were applied. In particular: (i) \textit{Gaussian Blur} considering different kernel sizes ($3x3$, $9x9$, $15x15$); (ii) \textit{Rotations} with degrees equal to 45$, 135$, 255$, 315$ degrees; (iii) \textit{Mirroring} along one axis (horizontal, vertical or both); (iv) \textit{Scaling} considering $-50\%$ and $+50\%$ w.r.t. the original image dimensions; (v) \textit{JPEG compression} with different quality factor values: $QF = \{50, 60, ..., 90\}$ and standard Quantization Tables. Many of these attacks are able to destroy several frequencies in Deepfake images making the extraction of a discriminative signature for each model more complex. Details of the noise-augmented version of the data is available at a dedicated site\footnote{\url{https://iplab.dmi.unict.it/mfs/Deepfakes/PaperCVPRW2022/}}. In the same way as before, a training-validation set split was performed (75\%-25\%) and the pre-trained Resnet-18 obtained before, for model classification on RAW images, was fine-tuned on this augmented dataset. The obtained training and validation plots are shown in Figure~\ref{fig:AccLossRESNETattacks} (a) and (b) and the overall classification results are reported in Figure~\ref{fig:AccLossRESNETattacks} in the form of a heat-map confusion matrix (c). In this case, the obtained average classification accuracy of $81.37\%$ demonstrated the robustness of the approach for the classification task on a-priori known models.



\subsection{Unknown model classification results}
\label{sub:resultsmetric}
Given a set of images ``in the wild", the task of Deepfake Model Recognition could not be carried out with techniques available in literature and with the classification model experimentally tested in the previous Section. Ideally, a first assessment could be performed using SOTA Deepfake detectors and architecture classifiers. Now, given a set of unknown models of a given architecture, the Model Recognition task could be carried out by exploiting a metric $L$. 

At first, metric $L$ was trained on a-priori known models ($M_{1-50}$ of the dataset) by means of $F$ (as defined and trained in the previous Section). The learnt metric $L$ was then employed in the tests. Three test sets were employed: (i) further $300 \times 50$ images generated by a-priori known models $M_{1-50}$; (ii) images generated by unknown models never used for any training operation $M_{51-100}$; (iii) the union of two previous sets and (iv) a noisy augmented version of (iii) with the noises described in Section \ref{subsec:RES}.

 All tests were carried out considering $20\%$ of each dataset as a reference and the remaining part was employed in a k-NN classification experiment employing the $L$ metric. 5-fold cross-validation was carried out and results, in terms of average accuracy, are reported in Table \ref{tab:metric_results}.

A further test was carried out on images belonging to the (i) set against images generated by a different architecture: StyleGAN \cite{karras2019style}. As presented in Table \ref{tab:metric_results}, an average accuracy of almost 100\%, demonstrates that this metric is able to maintain the correct distancing concept even between two different architectures, without any further training.
Finally, to introduce a preliminary concept of scalability, the learnt metric $L$ was further tested on images generated by two different pre-trained models of the StyleGAN-1 \cite{karras2019style} architecture ($30$ images for each model for a total of $60$). 

\begin{table}[]
\centering
\begin{adjustbox}{max width=\textwidth}
\begin{tabular}{c|l|l|l|}
\cline{2-4}
\multicolumn{1}{l|}{}                     & \multicolumn{1}{c|}{\textbf{1-NN}} & \multicolumn{1}{c|}{\textbf{3-NN}} & \multicolumn{1}{c|}{\textbf{5-NN}} \\ \hline
\multicolumn{1}{|c|}{\textbf{A-priori known}} &    95.08\%   &    95.74\%   & 95.74\%                                   \\ \hline
\multicolumn{1}{|c|}{\textbf{Unknown models}} & 92.31\%    &   92.55\%    & 94.33\%                                   \\ \hline
\multicolumn{1}{|c|}{\textbf{Known + unknown}} &  93.70\%   &   94.24\%    & 95.06\%                                   \\ \hline
\multicolumn{1}{|c|}{\textbf{Noise-augmented}} &    75.66\%         &   77.00\%    & 79.66\%                                   \\ \hline
\multicolumn{1}{|c|}{\textbf{ST1-M1 vs. StyleGAN2}}  &  99.98\%        &   99.98\%    & 99.98\%                                   \\ \hline
\multicolumn{1}{|c|}{\textbf{ST1-M1 vs. ST1-M2}}  &  84.50\%        &   88.20\%    & 92.00\%                                   \\ \hline
\end{tabular}
\end{adjustbox}
\caption{Results of the learnt metric $L$ by means of employing it in a k-NN classification test. Values reported are the average accuracies obtained on 5-fold cross validation tests employing 20\% of data as reference and the remaining for classification. ST1-M1 stands for the first model of StyleGAN1.}
\label{tab:metric_results}
\end{table}

\section{Discussion and Conclusion}
\label{sec:conclusion}

In this paper, a preliminary solution for Deepfake Model Recognition was presented. The underlying task, not yet completely addressed in the state-of-the-art, was described and attacked with an effective solution. 
The work proposed by Asnani et al~\cite{asnani2021reverse} is the most similar work, available in SOTA, compared to the proposed approach. Despite this, it is not possible to perform a comparison with~\cite{asnani2021reverse} because the proposed solution focuses on a more specific task: discriminating between different models created by the same GAN engine.
The images generated by the $100$ StyleGAN2 models were visually almost the same (Figure~\ref{fig:datasets}) and the main aim was to find features to discriminate them. Every new StyleGAN2 model has its own latent space that will be (even if in a ``minimal" way) different from the $100$ used in this paper. This happens because the training procedure will inevitably change the parameters of the neural architecture (so the latent space will be different). We are confident that the proposed method is scalable in any domain, because in general any other generative architecture, by its nature, will have a different latent representation since the architecture itself will be different. The ability to compare whether two images are generated by the same model,  sidesteps the problem of predicting/classifying a model out of potentially infinite possibilities.
The proposed methodology was demonstrated to be generalizable even for different architectures providing useful  insights for building a sort of feature dataset for each involved architecture similar to those PRNU-based for camera source identification methods. 
Finally, the proposed method was demonstrated to be effective in real-case scenarios. As a matter of fact, we widely tested the robustness of the technique in the wild: for instance IM platforms such as Whatsapp compress images with a quality factor equal to 50: in this context, our method performed with an accuracy of over $80\%$. 


\balance
{\small
\bibliographystyle{ieee_fullname}
\bibliography{main}

\begin{thebibliography}{10}\itemsep=-1pt

\bibitem{abdal2019image2stylegan}
Rameen Abdal, Yipeng Qin, and Peter Wonka.
\newblock Image2stylegan: How to embed images into the stylegan latent space?
\newblock In {\em Proceedings of the IEEE/CVF International Conference on
  Computer Vision}, pages 4432--4441, 2019.

\bibitem{agarwal2020detecting}
Shruti Agarwal, Hany Farid, Tarek El-Gaaly, and Ser-Nam Lim.
\newblock Detecting deep-fake videos from appearance and behavior.
\newblock In {\em 2020 IEEE International Workshop on Information Forensics and
  Security (WIFS)}, pages 1--6. IEEE, 2020.

\bibitem{asnani2021reverse}
Vishal Asnani, Xi Yin, Tal Hassner, and Xiaoming Liu.
\newblock Reverse engineering of generative models: Inferring model
  hyperparameters from generated images.
\newblock {\em arXiv preprint arXiv:2106.07873}, 2021.

\bibitem{battiato2016multimedia}
Sebastiano Battiato, Oliver Giudice, and Antonino Paratore.
\newblock Multimedia forensics: discovering the history of multimedia contents.
\newblock In {\em Proceedings of the 17th International Conference on Computer
  Systems and Technologies 2016}, pages 5--16, 2016.

\bibitem{bau2019seeing}
David Bau, Jun-Yan Zhu, Jonas Wulff, William Peebles, Hendrik Strobelt, Bolei
  Zhou, and Antonio Torralba.
\newblock {Seeing what a GAN Cannot Generate}.
\newblock In {\em Proceedings of the IEEE/CVF International Conference on
  Computer Vision}, pages 4502--4511, 2019.

\bibitem{creswell2018generative}
Antonia Creswell, Tom White, Vincent Dumoulin, Kai Arulkumaran, Biswa Sengupta,
  and Anil~A Bharath.
\newblock Generative adversarial networks: An overview.
\newblock {\em IEEE Signal Processing Magazine}, 35(1):53--65, 2018.

\bibitem{dang2020detection}
Hao Dang, Feng Liu, Joel Stehouwer, Xiaoming Liu, and Anil~K Jain.
\newblock On the detection of digital face manipulation.
\newblock In {\em Proceedings of the IEEE/CVF Conference on Computer Vision and
  Pattern Recognition}, pages 5781--5790, 2020.

\bibitem{durall2020watch}
Ricard Durall, Margret Keuper, and Janis Keuper.
\newblock Watch your up-convolution: {CNN} based generative deep neural
  networks are failing to reproduce spectral distributions.
\newblock In {\em Proceedings of the IEEE/CVF Conference on Computer Vision and
  Pattern Recognition}, pages 7890--7899, 2020.

\bibitem{durall2019unmasking}
R. Durall, M. Keuper, F. Pfreundt, and J. Keuper.
\newblock Unmasking deepfakes with simple features.
\newblock {\em arXiv preprint arXiv:1911.00686}, 2019.

\bibitem{farid2008digital}
Hany Farid.
\newblock Digital image ballistics from {JPEG} quantization: A followup study.
\newblock {\em Department of Computer Science, Dartmouth College, Tech. Rep.
  TR2008-638}, 2008.

\bibitem{frank2020leveraging}
Joel Frank, Thorsten Eisenhofer, Lea Sch{\"o}nherr, Asja Fischer, Dorothea
  Kolossa, and Thorsten Holz.
\newblock Leveraging frequency analysis for deep fake image recognition.
\newblock In {\em International Conference on Machine Learning}, pages
  3247--3258. PMLR, 2020.

\bibitem{giudice2021fighting}
Oliver Giudice, Luca Guarnera, and Sebastiano Battiato.
\newblock {Fighting Deepfakes by Detecting GAN DCT Anomalies}.
\newblock {\em Journal of Imaging}, 7(8), 2021.

\bibitem{goebel2021detection}
Michael Goebel, Lakshmanan Nataraj, Tejaswi Nanjundaswamy, Tajuddin~Manhar
  Mohammed, Shivkumar Chandrasekaran, and BS Manjunath.
\newblock Detection, attribution and localization of {GAN} generated images.
\newblock {\em Electronic Imaging}, 2021(4):276--1, 2021.

\bibitem{goodfellow2014generative}
Ian Goodfellow, Jean Pouget-Abadie, Mehdi Mirza, Bing Xu, David Warde-Farley,
  Sherjil Ozair, Aaron Courville, and Yoshua Bengio.
\newblock Generative adversarial nets.
\newblock In {\em Advances in Neural Information Processing Systems}, pages
  2672--2680, 2014.

\bibitem{guarnera2020deepfake}
Luca Guarnera, Oliver Giudice, and Sebastiano Battiato.
\newblock {DeepFake Detection by Analyzing Convolutional Traces}.
\newblock In {\em Proceedings of the IEEE/CVF Conference on Computer Vision and
  Pattern Recognition Workshops}, pages 666--667, 2020.

\bibitem{guarnera2020fighting}
Luca Guarnera, Oliver Giudice, and Sebastiano Battiato.
\newblock {Fighting Deepfake by Exposing the Convolutional Traces on Images}.
\newblock {\em IEEE Access}, 8:165085--165098, 2020.

\bibitem{hasan2019combating}
Haya~R Hasan and Khaled Salah.
\newblock Combating deepfake videos using blockchain and smart contracts.
\newblock {\em Ieee Access}, 7:41596--41606, 2019.

\bibitem{he2016deep}
Kaiming He, Xiangyu Zhang, Shaoqing Ren, and Jian Sun.
\newblock Deep residual learning for image recognition.
\newblock In {\em Proceedings of the IEEE Conference on Computer Vision and
  Pattern Recognition}, pages 770--778, 2016.

\bibitem{hulzebosch2020detecting}
Nils Hulzebosch, Sarah Ibrahimi, and Marcel Worring.
\newblock {Detecting {CNN}-Generated Facial Images in Real-World Scenarios}.
\newblock In {\em Proceedings of the IEEE/CVF Conference on Computer Vision and
  Pattern Recognition Workshops}, pages 642--643, 2020.

\bibitem{NEURIPS2020_8d30aa96}
Tero Karras, Miika Aittala, Janne Hellsten, Samuli Laine, Jaakko Lehtinen, and
  Timo Aila.
\newblock Training generative adversarial networks with limited data.
\newblock In H. Larochelle, M. Ranzato, R. Hadsell, M.~F. Balcan, and H. Lin,
  editors, {\em Advances in Neural Information Processing Systems}, volume~33,
  pages 12104--12114. Curran Associates, Inc., 2020.

\bibitem{karras2019style}
Tero Karras, Samuli Laine, and Timo Aila.
\newblock A style-based generator architecture for generative adversarial
  networks.
\newblock In {\em Proceedings of the IEEE/CVF Conference on Computer Vision and
  Pattern Recognition}, pages 4401--4410, 2019.

\bibitem{karras2020analyzing}
Tero Karras, Samuli Laine, Miika Aittala, Janne Hellsten, Jaakko Lehtinen, and
  Timo Aila.
\newblock Analyzing and improving the image quality of stylegan.
\newblock In {\em Proceedings of the IEEE/CVF Conference on Computer Vision and
  Pattern Recognition}, pages 8110--8119, 2020.

\bibitem{liu2012metric}
Eric~Yi Liu, Zhishan Guo, Xiang Zhang, Vladimir Jojic, and Wei Wang.
\newblock Metric learning from relative comparisons by minimizing squared
  residual.
\newblock In {\em 2012 IEEE 12th International Conference on Data Mining},
  pages 978--983. IEEE, 2012.

\bibitem{marra2019gans}
Francesco Marra, Diego Gragnaniello, Luisa Verdoliva, and Giovanni Poggi.
\newblock Do {GANs} leave artificial fingerprints?
\newblock In {\em 2019 IEEE Conference on Multimedia Information Processing and
  Retrieval (MIPR)}, pages 506--511. IEEE, 2019.

\bibitem{masood2021deepfakes}
Momina Masood, Marriam Nawaz, Khalid~Mahmood Malik, Ali Javed, and Aun Irtaza.
\newblock Deepfakes generation and detection: State-of-the-art, open
  challenges, countermeasures, and way forward.
\newblock {\em arXiv preprint arXiv:2103.00484}, 2021.

\bibitem{mirsky2021creation}
Yisroel Mirsky and Wenke Lee.
\newblock The creation and detection of deepfakes: A survey.
\newblock {\em ACM Computing Surveys (CSUR)}, 54(1):1--41, 2021.

\bibitem{ong2021protecting}
Ding~Sheng Ong, Chee~Seng Chan, Kam~Woh Ng, Lixin Fan, and Qiang Yang.
\newblock Protecting intellectual property of generative adversarial networks
  from ambiguity attacks.
\newblock In {\em Proceedings of the IEEE/CVF Conference on Computer Vision and
  Pattern Recognition}, pages 3630--3639, 2021.

\bibitem{piva2013overview}
Alessandro Piva.
\newblock An overview on image forensics.
\newblock {\em International Scholarly Research Notices}, 2013, 2013.

\bibitem{schwettmann2021toward}
Sarah Schwettmann, Evan Hernandez, David Bau, Samuel Klein, Jacob Andreas, and
  Antonio Torralba.
\newblock {Toward a Visual Concept Vocabulary for GAN Latent Space}.
\newblock In {\em Proceedings of the IEEE/CVF International Conference on
  Computer Vision}, pages 6804--6812, 2021.

\bibitem{selvaraju2017grad}
Ramprasaath~R Selvaraju, Michael Cogswell, Abhishek Das, Ramakrishna Vedantam,
  Devi Parikh, and Dhruv Batra.
\newblock Grad-cam: Visual explanations from deep networks via gradient-based
  localization.
\newblock In {\em Proceedings of the IEEE international conference on computer
  vision}, pages 618--626, 2017.

\bibitem{shen2020interpreting}
Yujun Shen, Jinjin Gu, Xiaoou Tang, and Bolei Zhou.
\newblock {Interpreting the Latent Space of GANs for Semantic Face Editing}.
\newblock In {\em Proceedings of the IEEE/CVF Conference on Computer Vision and
  Pattern Recognition}, pages 9243--9252, 2020.

\bibitem{swathi2021deepfake}
P Swathi and Saritha Sk.
\newblock Deepfake creation and detection: A survey.
\newblock In {\em 2021 Third International Conference on Inventive Research in
  Computing Applications (ICIRCA)}, pages 584--588. IEEE, 2021.

\bibitem{tolosana2020deepfakes}
Ruben Tolosana, Ruben Vera-Rodriguez, Julian Fierrez, Aythami Morales, and
  Javier Ortega-Garcia.
\newblock Deepfakes and beyond: A survey of face manipulation and fake
  detection.
\newblock {\em arXiv preprint arXiv:2001.00179}, 2020.

\bibitem{uchida2017embedding}
Yusuke Uchida, Yuki Nagai, Shigeyuki Sakazawa, and Shin'ichi Satoh.
\newblock Embedding watermarks into deep neural networks.
\newblock In {\em Proceedings of the 2017 ACM on International Conference on
  Multimedia Retrieval}, pages 269--277, 2017.

\bibitem{verdoliva2020media}
Luisa Verdoliva.
\newblock Media forensics and deepfakes: an overview.
\newblock {\em IEEE Journal of Selected Topics in Signal Processing},
  14(5):910--932, 2020.

\bibitem{vukotic2018deep}
Vedran Vukoti{\'c}, Vivien Chappelier, and Teddy Furon.
\newblock Are deep neural networks good for blind image watermarking?
\newblock In {\em 2018 IEEE International Workshop on Information Forensics and
  Security (WIFS)}, pages 1--7. IEEE, 2018.

\bibitem{wang2019fakespotter}
R. Wang, L. Ma, F. Juefei-Xu, X. Xie, J. Wang, and Y. Liu.
\newblock Fakespotter: A simple baseline for spotting ai-synthesized fake
  faces.
\newblock {\em arXiv preprint arXiv:1909.06122}, 2019.

\bibitem{wang2020cnn}
Sheng-Yu Wang, Oliver Wang, Richard Zhang, Andrew Owens, and Alexei~A Efros.
\newblock {CNN-generated images are surprisingly easy to spot... for now}.
\newblock In {\em Proceedings of the IEEE/CVF Conference on Computer Vision and
  Pattern Recognition}, pages 8695--8704, 2020.

\bibitem{yu2019attributing}
Ning Yu, Larry~S Davis, and Mario Fritz.
\newblock Attributing fake images to {GANs}: Learning and analyzing {GAN}
  fingerprints.
\newblock In {\em Proceedings of the IEEE/CVF International Conference on
  Computer Vision}, pages 7556--7566, 2019.

\bibitem{yu2020responsible}
Ning Yu, Vladislav Skripniuk, Dingfan Chen, Larry Davis, and Mario Fritz.
\newblock Responsible disclosure of generative models using scalable
  fingerprinting.
\newblock {\em arXiv preprint arXiv:2012.08726}, 2020.

\bibitem{zhang2020not}
Baiwu Zhang, Jin~Peng Zhou, Ilia Shumailov, and Nicolas Papernot.
\newblock Not my deepfake: Towards plausible deniability for machine-generated
  media.
\newblock {\em arXiv e-prints}, pages arXiv--2008, 2020.

\bibitem{zhang2018protecting}
Jialong Zhang, Zhongshu Gu, Jiyong Jang, Hui Wu, Marc~Ph Stoecklin, Heqing
  Huang, and Ian Molloy.
\newblock Protecting intellectual property of deep neural networks with
  watermarking.
\newblock In {\em Proceedings of the 2018 on Asia Conference on Computer and
  Communications Security}, pages 159--172, 2018.

\bibitem{zhang2019invisible}
Ru Zhang, Shiqi Dong, and Jianyi Liu.
\newblock Invisible steganography via generative adversarial networks.
\newblock {\em Multimedia tools and applications}, 78(7):8559--8575, 2019.

\bibitem{zhang2019detecting}
Xu Zhang, Svebor Karaman, and Shih-Fu Chang.
\newblock Detecting and simulating artifacts in {GAN} fake images.
\newblock In {\em 2019 IEEE International Workshop on Information Forensics and
  Security (WIFS)}, pages 1--6. IEEE, 2019.

\end{thebibliography}
}

\end{document}